\documentclass[10pt,twocolumn,letterpaper]{article}

\usepackage[pagenumbers]{cvpr} 

\definecolor{cvprblue}{rgb}{0.21,0.49,0.74}
\usepackage[pagebackref,breaklinks,colorlinks,allcolors=cvprblue]{hyperref}

\usepackage{algorithm}
\usepackage{algpseudocode} 
\usepackage{color}
\usepackage{xcolor}  
\definecolor{highlight}{RGB}{200,0,0}
\usepackage{listings}
\usepackage{newfloat}

\usepackage{subcaption}
\usepackage{enumitem} 
\usepackage{multirow} 
\usepackage{makecell} 
\usepackage{pifont} 
\usepackage{amssymb} 
\usepackage{amsmath}
\usepackage{booktabs}

\usepackage{microtype}
\usepackage{tabularx}
\usepackage{xspace}
\usepackage{mparhack}

\def\paperID{13147} 
\def\confName{CVPR}
\def\confYear{2026}

\title{KernelDNA: Dynamic Kernel Sharing via Decoupled Naive Adapters}

\author{Haiduo Huang\thanks{{\tt\small huanghd@stu.xjtu.edu.cn}},\; Yadong Zhang,\; Yinghui Xu,\; Pengju Ren\\
Institute of Artificial Intelligence and Robotics,\; Xi'an Jiaotong University\\
}

\begin{document}
\maketitle

\begin{abstract}
    Dynamic convolution enhances model capacity by adaptively combining multiple kernels, yet faces critical trade-offs: prior works either (1) incur significant parameter overhead by scaling kernel numbers linearly, (2) compromise inference speed through complex kernel interactions, or (3) struggle to jointly optimize dynamic attention and static kernels. We observe that pre-trained Convolutional Neural Networks (CNNs) exhibit inter-layer redundancy akin to that in Large Language Models (LLMs). Specifically, dense convolutional layers can be efficiently replaced by derived ``child" layers generated from a shared ``parent" convolutional kernel through an adapter. To address these limitations and implement the weight-sharing mechanism, we propose a lightweight convolution kernel plug-in, named KernelDNA. It decouples kernel adaptation into input-dependent dynamic routing and pre-trained static modulation, ensuring both parameter efficiency and hardware-friendly inference. Unlike existing dynamic convolutions that expand parameters via multi-kernel ensembles, our method leverages cross-layer weight sharing and adapter-based modulation, enabling dynamic kernel specialization without altering the standard convolution structure. This design preserves the native computational efficiency of standard convolutions while enhancing representation power through input-adaptive kernel adjustments. Experiments on image classification and dense prediction tasks demonstrate that KernelDNA achieves state-of-the-art accuracy-efficiency balance among dynamic convolution variants.
\end{abstract}

\section{Introduction}
\label{sec:intro}
Convolutional Neural Networks~\cite{he2016deep,sandler2018mobilenetv2,liu2022convnet} have long been built on static convolutional kernels within each layer. While these static kernels enable computationally efficient inference, their fixed, input-agnostic nature inherently limits representation flexibility. Dynamic convolution~\cite{Yang2019,Chen2020} attempts to address this by adaptively modulating kernels based on input content, but it inevitably scales parameters linearly with the number of kernels, introducing prohibitive memory costs that hinder real-world deployment. Existing model compression techniques, such as pruning~\cite{Li2016}, quantization~\cite{Esser2019}, and knowledge distillation~\cite{Gou2021}, aim to alleviate parameter growth but require complex multi-stage pipelines: they often depend on fine-tuning pre-trained models with carefully tuned hyperparameters. In contrast, weight-sharing~\cite{Wang2024a,Cao2024} strategies in LLMs demonstrate that parameter redundancy can be exploited for drastic compression without performance loss. Inspired by this, we propose integrating dynamic neural networks' adaptive capacity with LLM-like weight sharing, unlocking the potential of CNNs through a novel parameter-efficient dynamic modulation framework.

\begin{figure}[ht]
    \centering
    \includegraphics[width=0.47\textwidth]{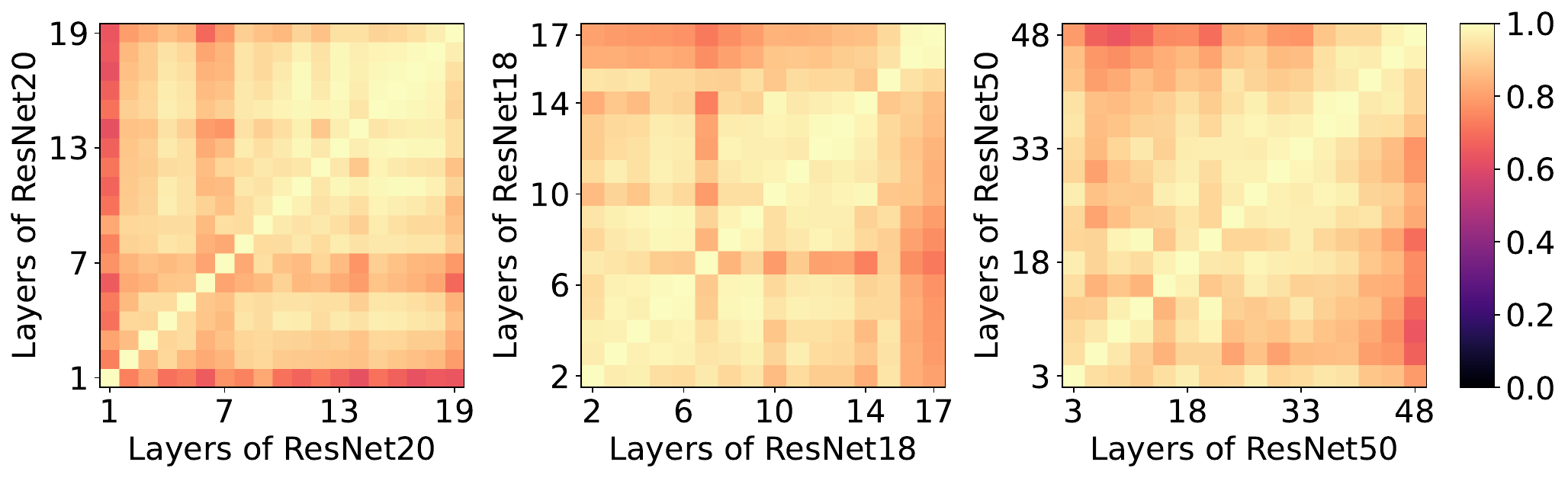}
    \caption{Linear CKA across layers of different models reveals a consistent grid pattern, derived from models trained on their original architectures. Only the results for Conv3$\times$3 are presented. Except for the last two layers and a few intermediate layers, the similarity between different layers is notably high.}
    \label{fig:resnets_CKA}
    \vspace{-0.3cm}
\end{figure}

\begin{figure}[ht]
    \centering
    \includegraphics[width=0.4\textwidth]{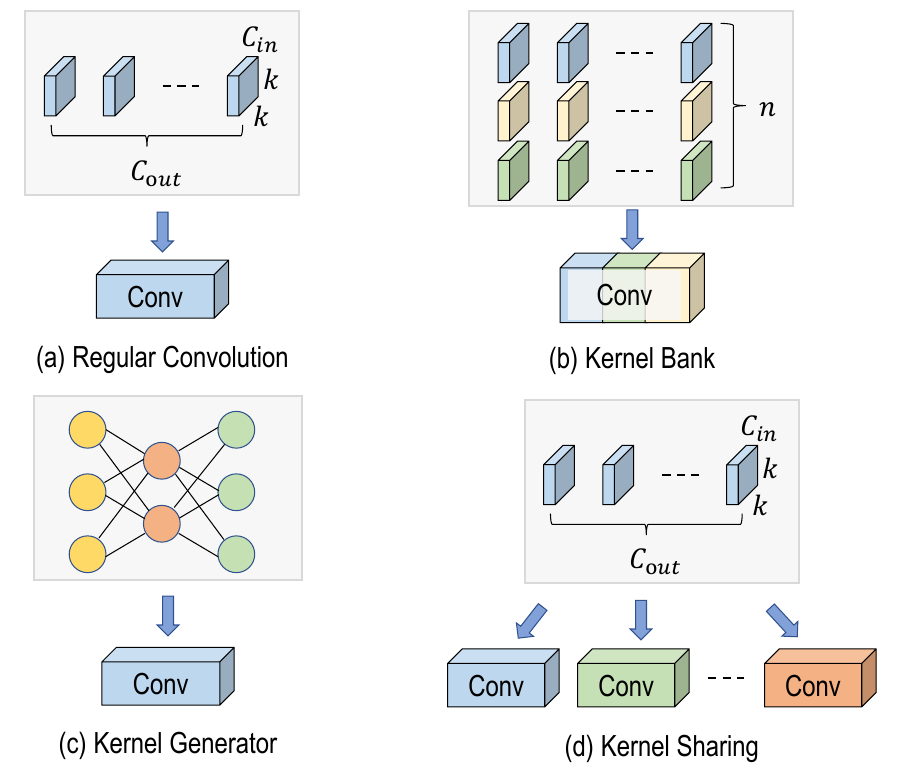}
    \caption{Prior dynamic convolution methods are mainly classified into two categories: \ie, (b) and (c). (a) Standard convolution with fixed kernels. (b) Kernel Pool-based~\cite{Yang2019,Chen2020,Li2022,Li2024}, which maintains a pool of convolutional kernels trained alongside other model parameters and are fixed after training, and (c) Neural Generator-based~\cite{zhou2021decoupled,Li2021,he2023sd,Chen2025b}, which employs a generator network or hybrid attention mechanisms to directly synthesize convolutional kernels from the input. (d) Our proposed KernelDNA approach with shared parent kernels and adapter-modulated child layers.}
    \label{fig:diff_Dynamic_Conv}
    \vspace{-0.3cm}
\end{figure}

To validate the inherent inter-layer redundancy in conventional CNNs, we visualize the similarity between convolutional kernels across layers using the Linear Centered Kernel Alignment (CKA) index~\cite{Kornblith2019}. As shown in Figure~\ref{fig:resnets_CKA}, except for a few layers that exhibit low similarity with others, the majority of layers show high similarity. This observation suggests that compression or transformation can be applied to enable similar layers to share a base layer, thereby reducing the model's parameters without compromising the architecture. Drawing inspiration from Epigenetics~\cite{waddington1942epigenotype}—where phenotypic traits are regulated through non-heritable modifications (\eg, DNA methylation) without altering the underlying genetic code—we design a lightweight adapter module. This module decouples input-dependent dynamic channel attention from kernel-centric spatial modulation, enabling a shared ``parent" kernel within a block or stage to generate derived ``child" layers via adapter-based transformations. By replacing dense convolutional layers with these adapted child layers, we achieve substantial parameter reduction while preserving performance. Figure~\ref{fig:diff_Dynamic_Conv} illustrates the distinction between the kernel sharing mechanism in our method and mainstream dynamic convolution approaches. 

Both previous paradigms (\ie,Figure~\ref{fig:diff_Dynamic_Conv} (b) and (c)) share critical limitations: they introduce computationally intensive operations (\eg, dynamic kernel assembly or generation) that degrade inference speed and often cause parameter inflation. In contrast, our kernel-sharing strategy not only reduces parameters but also preserves the efficient inference advantage of traditional convolution by reusing and modulating shared base kernels across layers. Notably, parameter sharing introduces implicit regularization~\cite{Lan2019}, mitigating overfitting risks and even enhancing generalization.

To ensure hardware-friendly efficiency, our adapter integrates three enhanced attention mechanisms: Channel Attention, Filter Attention, and Spatial Attention. These mechanisms allow each convolutional layer to retain distinct characteristics while benefiting from cross-layer captured through shared parameters. Termed {\bf KernelDNA}, this plug-and-play module functions as an epigenetic ``marker", dynamically tailoring kernel behavior without altering the base parameter structure. It is universally applicable to any standard convolution operator, including regular convolutions, depthwise convolutions, and 1$\times$1 convolutions.

Extensive experiments on ImageNet-1K and downstream tasks demonstrate that KernelDNA surpasses prior dynamic convolution methods in accuracy while preserving near-original inference speeds (\eg, 7,683 fps vs. ResNet18's 8,463 fps, 90.8\% throughput retention). Compared to SOTA variants like FDConv (11.75M params), KernelDNA achieves 1.27$\times$ parameter reduction (9.24M for ResNet18) with higher accuracy (74.23\% vs. 71.25\% Top-1). On lightweight models like MobileNetV2, KernelDNA attains 75.51\% Top-1 accuracy (vs. KernelWarehouse's 74.68\%) using only 3.14M parameters (vs. KernelWarehouse's 5.17M), while maintaining 99\% of the base model's throughput (6,260 fps vs. 6,323 fps). These results establish a new Pareto frontier, where KernelDNA uniquely balances parameter efficiency, hardware compatibility, and adaptive performance across diverse architectures.

\section{Related Work}
\noindent
{\bf Dynamic Convolution.}
Dynamic convolution aims to enhance the flexibility of CNNs by allowing kernel weights to adapt to input data, breaking the limitations of static convolutions. Early methods like CondConv~\cite{Yang2019} use input-dependent routing to combine multiple fixed kernels, while DY-CNNs~\cite{Chen2020} leverage channel attention to control dynamic aggregation. Later, approaches such as DCD~\cite{Li2021} and DD~\cite{Zhou2021} introduce low-rank decompositions or separate channel and spatial modulation, enabling finer adaptive behavior with reduced overhead. More recent designs, including ODConv~\cite{Li2022} and SD-Conv~\cite{He2023}, further generalize adaptiveness to spatial sizes or sparsify kernels for efficiency, but often at the cost of increased complexity, memory, or unstable latency. KernelWarehouse~\cite{Li2024} assembles kernels from a large bank of fine-grained components, and FDConv~\cite{Chen2025b} adapts kernels in the frequency domain. However, despite improvements, most dynamic convolution methods either introduce heavy memory and latency overhead due to complex kernel synthesis at inference, or trade off accuracy for efficiency. Our KernelDNA addresses these shortcomings by adopting cross-layer weight sharing and lightweight adapters, delivering effective dynamic modulation with minimal inference cost and robust throughput.

\noindent
{\bf Weight Sharing.}
Weight sharing has been widely investigated in NLP, starting with ALBERT~\cite{Lan2019} for parameter-efficient transformers by tying parameters across layers. These ideas inspired similar work in vision models, such as parameter multiplexing in MiniViT~\cite{Zhang2022}, low-rank decomposition in MPOBERT~\cite{Liu2023a}, RL-based dynamic layer reuse~\cite{Hay2024}, and SVD-based basis sharing~\cite{Wang2024a}. Further, Relaxed Recursive Transformers~\cite{Bae2024} integrate LoRA-like adapters for effective sharing. Despite their success in transformer-based models, the application of deep weight sharing in CNNs, particularly under dynamic convolution settings, remains relatively unexplored. Our work is among the first to systematically integrate weight sharing and dynamic convolution, achieving both high parameter efficiency and fast inference.

\noindent
{\bf Visual Attention Modules.}
Visual attention mechanisms are now integral to improving CNN feature representations. SE-Net~\cite{hu2018squeeze} introduced channel-wise attention via global pooling and fully connected layers, while CBAM~\cite{woo2018cbam} further augments channel attention with spatial attention for joint modulation. ECA-Net~\cite{wang2020eca} simplifies channel interaction using efficient local 1D convolutions, and CGC~\cite{lin2020context} employs context gating to integrate input and global context for kernel modulation. Different from these, KernelDNA separates dynamic channel attention from static filter and spatial attention, employing a lightweight adapter that minimizes computational costs, thereby improving both adaptability and efficiency in practical CNN deployments.

\section{Preliminaries}
\noindent {\bf Standard Convolution.}  
Given an input feature map {\small \( \mathbf{X} \in \mathbb{R}^{c \times h \times w} \)} with {\small \( c \)} channels and spatial dimensions {\small \( h \times w \)}, the standard convolution operation applies a fixed filter {\small \( \mathbf{W} \in \mathbb{R}^{c' \times c \times k \times k} \)} to produce an output feature map {\small \( \mathbf{Y} \in \mathbb{R}^{c' \times h' \times w'} \)}. At each output spatial location {\small \( (i, j) \)}, the convolution is computed as:  
{\small
\begin{equation}
\mathbf{Y}_{(m,i,j)} = \sum_{n,p,q} \mathbf{W}_{(m,n,p,q)} \cdot \mathbf{X}_{(n,i+p-\lfloor k/2 \rfloor, j+q-\lfloor k/2 \rfloor)} + \mathbf{b}_{m},  
\end{equation}
}where {\small \( m \in {\{0,...,c'-1\}} \)} is the output channel index, {\small \( n \in {\{0,...,c-1\}} \)} is the input channel index, {\small \( p,q \in {\{0,...,k-1\}} \)} are the spatial indices, and {\small \( \mathbf{b} \in \mathbb{R}^{c'} \)} is the bias term. The spatial indices are adjusted with appropriate padding to ensure valid indexing. The filter {\small \( \mathbf{W} \)} is shared across all spatial locations, making the operation computationally efficient but input-agnostic. Here, the channel dimension {\small \( n \)} is explicitly summed over, ensuring proper alignment between the input channels and the kernel's input channel dimension.

\noindent {\bf Conventional Dynamic Convolution.}  
Dynamic convolution improves efficiency by aggregating a set of \( n \) static kernels {\small \( \{\mathbf{W}_1, \dots, \mathbf{W}_n\} \)} where each {\small \( \mathbf{W}_i \in \mathbb{R}^{c' \times c \times k \times k} \)} using input-dependent attention weights {\small \( \pi_i(\mathbf{X}) \in \mathbb{R} \)}:  
{\small
\begin{equation}
\mathbf{W} = \sum_{i=1}^n \pi_i(\mathbf{X}) \cdot \mathbf{W}_i,  
\end{equation}
}where {\small \( \pi_i(\mathbf{X}) \)} is a scalar weight computed via a lightweight attention mechanism such that {\small \( \sum_{i=1}^n \pi_i(\mathbf{X}) = 1 \)}. The aggregated kernel {\small \( \mathbf{W} \in \mathbb{R}^{c' \times c \times k \times k} \)} maintains the same dimensions as each individual kernel {\small \( \mathbf{W}_i \)} and is then applied using standard convolution.

Although dynamic convolution enhances model capacity by introducing multiple static kernels, it inevitably incurs additional parameters and computational costs, negatively impacting inference efficiency. To address these issues, our proposed KernelDNA tackles the parameter inefficiency of dynamic convolution through a {\bf layer-wise weight-sharing} mechanism. By integrating lightweight visual adapters, KernelDNA enables shared kernels to retain unique characteristics, effectively replacing dense convolutions in their original positions. The entire process is highly efficient, with minimal impact on throughput.

\section{Methodology}
\subsection{Dynamic Kernel Sharing}
\begin{figure}[ht]
    \centering
    \includegraphics[width=0.3\textwidth]{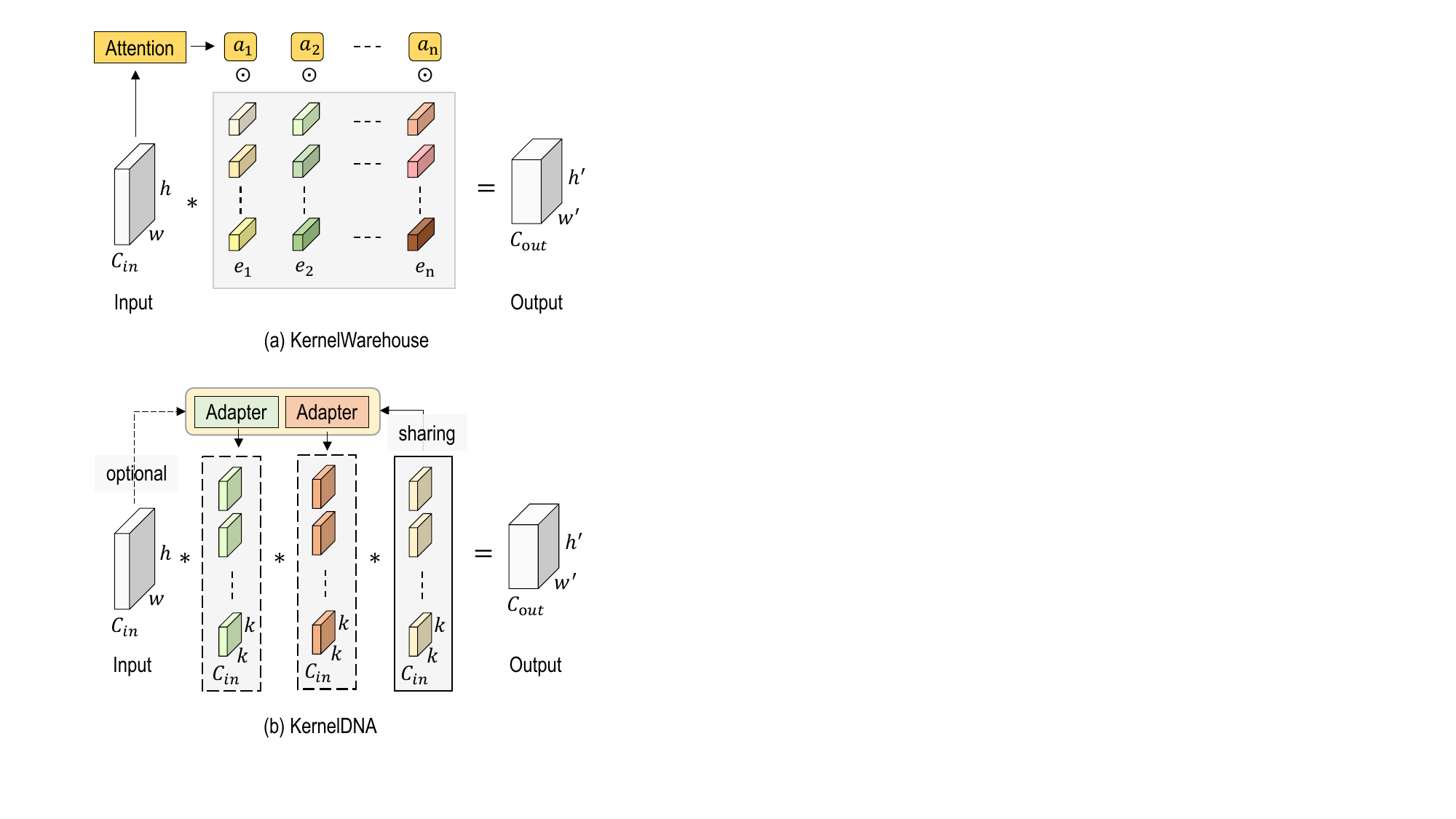}
    \caption{Comparison of dynamic convolution, where $*$ denotes the convolution operation.}
    \label{fig:KernelDNA}
    \vspace{-0.3cm}
\end{figure}

Recent works have aimed to enhance the flexibility of dynamic convolution by increasing kernel diversity or granularity. For instance, KernelWarehouse~\cite{Li2024} decomposes each convolutional kernel into hundreds of fine-grained cell-wise components {\small $e_n$}, building a global bank of reusable kernel elements. During inference, it assembles full kernels on-the-fly using input-driven, contrast-based attention, then applies them to feature maps. FDConv~\cite{Chen2025b} instead leverages Fourier-based kernels for improved adaptability but introduces significant memory consumption and additional computational steps. Although KernelWarehouse advertises cross-layer kernel sharing, it still relies on aggregating candidate static kernels via attention—much like classic kernel bank approaches, just at finer granularity (see Figure~\ref{fig:KernelDNA}~(a)). And this strategy calculates attention over many components of kernels and synthesizes full kernels at runtime, leading to considerable {\bf computational overhead}. Such operations demand excessive memory access and are unfriendly to hardware, resulting in much slower inference than standard convolutions and limiting deployment in latency-sensitive applications.

In contrast to KernelWarehouse's fine-grained decomposition and reassembly, our KernelDNA preserves the structure of standard convolutions while introducing {\bf layer-wise parameter sharing}. Instead of complex computations, KernelDNA employs lightweight adapters to generate layer-specific kernels from shared base parameters, retaining the inference efficiency of standard convolutions. This design enables seamless replacement of traditional convolutions, as depicted in Figure~\ref{fig:KernelDNA}~(b), where the solid box representations a static convolutional kernel, which serves as the parent layer in KernelDNA. The dashed boxes denote child layers derived from the parent layer through adapter transformations. These child layers do not have their own weights, which are dynamically generated during inference to perform convolution operations.

Furthermore, our adapter integrates both static attention (determined during training) and input-dependent dynamic attention (computed at inference), decoupling purely dynamic attention mechanisms of previous works. This dual-pathway design enables flexible adaptation to different deployment scenarios: dynamic attention can be disabled for minimal latency with only a slight accuracy trade-off, or enabled to boost performance at the cost of marginal additional computational overhead. Thus, KernelDNA can serve as a plug-and-play replacement for conventional convolution layers, balancing efficiency and adaptability without compromising hardware compatibility.

\subsection{Light-weight Modulation Adapter}
\begin{figure}[ht]
  \centering
  \includegraphics[width=0.36\textwidth]{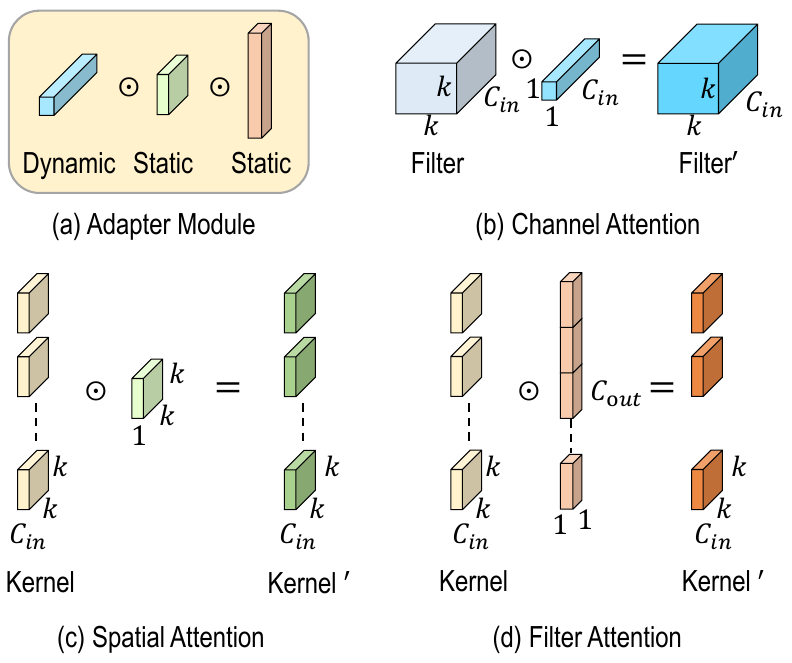}
  \caption{Light-weight Input-dependent Adapter. Note that the channel attention is shared across all filters, and can be applied to the input feature map equivalently, which avoids the batch expansion of the kernel tensors.}
  \label{fig:Adapter}
  \vspace{-0.3cm}
\end{figure}

To efficiently transform base (parent) kernels into shared (child) kernels, we design a lightweight adapter that incorporates three attention mechanisms to modulate kernel properties across different dimensions: input channel, output channel, and spatial interactions. These mechanisms are termed Channel Attention, Filter Attention, and Spatial Attention, respectively. As illustrated in Figure~\ref{fig:Adapter}, Channel Attention is dynamically computed based on input features, while Filter Attention and Spatial Attention are statically stored as model parameters. The static mechanisms can be efficiently implemented in parallel, reducing the computational overhead and parameter count compared to purely dynamic attention mechanisms such as ODConv.

Notably, the static attention weights can be fused with the parent kernel during initialization to precompute child kernels, trading memory for computational efficiency. Alternatively, child kernels can be dynamically generated at inference time with a minor latency penalty. The mathematical formulation is defined as:
{\small
\begin{equation}
    \mathbf{\widetilde{W}}^{(l)} = \mathbf{W} \odot \underbrace{\overbrace{\beta_c^{(l)}(\mathbf{X}^{(l)})}^{{\bf channel\,modulate}}}_{{\bf dynamic}} \odot \underbrace{\overbrace{(1+\alpha_f^{(l)})}^{{\bf filter\,modulate}}  \odot \overbrace{(1+\alpha_s^{(l)})}^{{\bf spatial\,modulate}}}_{{\bf static}}
\end{equation}
}where {\small $\mathbf{W} \in \mathbb{R}^{c' \times c \times k \times k}$} is the base kernel shared across multiple identical convolutional layers, {\small $\mathbf{\widetilde{W}}^{(l)} \in \mathbb{R}^{c' \times c \times k \times k}$} is the modulated kernel for {\small $l$}-th layer, and {\small $\beta_c^{(l)}$, $\alpha_f^{(l)}$, and $\alpha_s^{(l)}$} are three components of {\small $l$}-th adapter. The {\small $\odot$} represents element-wise multiplication with broadcasting.

\noindent {\bf Channel Attention:} In fact, applying Channel Attention to convolutional kernels is equivalent to applying it to feature maps. We adopt the latter approach for efficient implementation. Our implementation resembles SE-Net~\cite{hu2018squeeze} but incorporates a Batch Normalization (BN) layer after the first fully connected (FC) layer. The default channel reduction ratio is set to 4. The detailed implementation process is as follows:
{\small
\begin{equation}
    \beta_c(\mathbf{X}) = \sigma\left\{ \mathbf{FC_2} \left[ \operatorname{ReLU} \left( \operatorname{BN}\big( \mathbf{FC_1} \left( \operatorname{AvgPool}(\mathbf{X}) \right) \big) \right) \right] \right\}
\end{equation}
}where the {\small $\sigma$} denotes the sigmoid function, {\small $\operatorname{AvgPool}$} denotes the global average pooling operation.

\noindent {\bf Spatial Attention and Filter Attention:} Compared to the implementation of Channel Attention, Spatial Attention and Filter Attention are simpler. They are implemented as learnable tensors with dimensions {\small $\alpha_s^{(l)} \in \mathbb{R}^{1 \times 1 \times k \times k}$} and {\small $\alpha_f^{(l)} \in \mathbb{R}^{c' \times 1 \times 1 \times 1}$}, and are stored as static parameters of the adapter. Once trained, these parameters remain fixed and can be fused with the parent kernel during initialization for inference. The Spatial Attention {\small $\alpha_s^{(l)}$} modulates the spatial distribution of convolutional kernels, enhancing sensitivity to critical regions (\eg, edges or textures)~\cite{woo2018cbam}. The Filter Attention {\small $\alpha_f^{(l)}$} adaptively scales output channels, amplifying discriminative features while suppressing noise. The design enables input-aware kernel modulation while retaining the computational benefits of standard convolutions (Figure~\ref{fig:Adapter}(b-d)).

\begin{algorithm*}[ht]\small
    \caption{Comparison of Purely Dynamic Attention vs. Decoupled Attention Pseudocode, PyTorch-like}
    \label{alg:pseudocode}
    \definecolor{codeblue}{rgb}{0.25,0.5,0.5}
    \definecolor{codekw}{rgb}{0.85, 0.18, 0.50}
    \lstset{
      backgroundcolor=\color{white},
      basicstyle=\fontsize{7.5pt}{7.5pt}\ttfamily\selectfont,
      columns=fullflexible,
      breaklines=true,
      captionpos=b,
      commentstyle=\fontsize{7.5pt}{7.5pt}\color{codeblue},
      keywordstyle=\fontsize{7.5pt}{7.5pt}\color{codekw},
    }
\begin{lstlisting}[language=python]
    # Input x: [B, C_in, H, W], Output y: [B, C_out, H, W]

    # Purely dynamic attention for ODConv
        weight = nn.Parameter(torch.randn(N, C_out, C_in, k, k))
        kernel_attn, filter_attn, channel_attn, spatial_attn = Multi-dimensional_Attention(x)

        # filter_attn: [B, C_out, 1, 1]  spatial_attn: [B, 1, 1, 1, k, k]
        # channel_attn: [B, C_in, 1, 1]  kernel_attn:  [B, N, 1, 1, 1, 1] 

        # Batch-expanded weights (Memory intensive)
        weight = weight.unsqueeze(dim=0) * kernel_attn * spatial_attn # [B, N, C_out, C_in, k, k] 
        weight = torch.sum(weight, dim=1).view(-1, C_in, k, k) # [B*C_out, C_in, k, k]

        x = (x * channel_attn).view(1, B*C_in, H, W) # Requiring Batch-grouped convolution
        y = F.conv2d(x, weight, groups=B).view(B, C_out, H, W) * filter_attn # Apply filter_attn sequentially! 

    # Decoupled attention for our KernelDNA
        weight = parent_conv.weight # [C_out, C_in, k, k], sharing across multiple identical conv layers
        filter_attn, spatial_attn = nn.Parameter(torch.zeros(C_out,1,1,1)), nn.Parameter(torch.zeros(1,1,k,k))
        channel_attn = SE-like_Attention(x) # [B, C_in, 1, 1]
        
        weight = weight * (1 + filter_attn) * (1 + spatial_attn) # [C_out, C_in, k, k] (No batch expansion, and these two operations can be applied in parallel with channel_attn below!)

        x = x * channel_attn    # [B, C_in, H, W]
        y = F.conv2d(x, weight) # [B, C_out, H, W]
\end{lstlisting}

\begin{algorithmic}\small
\State {\bf Key Differences:}
\State \begin{tabular}{p{8.2cm}p{8cm}}
\textcolor{highlight}{Purely Dynamic Attention (ODConv)} & \textcolor{highlight}{Decoupled Attention (KernelDNA)} \\
\hline
- Weight tensor grows with batch size ($B\times N\times C_{out}\times C_{in}\times k\times k$) & - Fixed weight size ($C_{out}\times C_{in}\times k\times k$) \\
- Requires memory-intensive view/reshape operations &- No tensor reshaping needed \\
- $O(B\cdot N\cdot C_{out}\cdot C_{in}\cdot k^2)$ parameters per batch &- $O(C_{out}\cdot C_{in}\cdot k^2)$ fixed parameters \\
- Complex grouped convolution &- Standard convolution \\
\end{tabular}
\end{algorithmic}
\end{algorithm*}

Through analysis, we observe that previous works' purely dynamic computation of spatial attention and filter attention introduces significant storage overhead. By decoupling these components and applying them in parallel, we achieve both efficiency and expressiveness. This decoupled design inspired the name ``DNA" for our adapter. The pseudocode implementation is shown in Algorithm~\ref{alg:pseudocode}.

\section{Experiments}
In this section, to validate the effectiveness of our proposed KernelDNA, we primarily evaluate image classification accuracy on the ImageNet-1K~\cite{russakovsky2015imagenet} dataset and downstream task performance (object detection and image segmentation) on the MS-COCO~\cite{lin2014microsoft} dataset. For simplicity, in the following tables, ``KW" denotes the KernelWarehouse method, while ``KDNA" represents our proposed KernelDNA. All experiments were conducted on a system equipped with an Nvidia GeForce RTX 4090 GPU and an AMD EPYC 7H12 64-Core Processor CPU. To ensure consistency, latency is reported with an input batch size of 1 and single-threaded CPU execution, while throughput is measured with a batch size of 128. To minimize variance, all results are averaged over 1,000 iterations. The models are trained from scratch, consistent with all the other compared methods.

\subsection{Image Classification on ImageNet-1K}
\noindent {\bf Experiment Setup:}
To benchmark against SOTA methods, we evaluate our approach on three representative architectures: standard ResNet18/50~\cite{he2016deep}, advanced ConvNeXt-Tiny~\cite{liu2022convnet}, and compact MobileNetV2~\cite{sandler2018mobilenetv2}. All experiments are implemented using publicly available PyTorch official codebases, with training strategies aligned with those of KernelWarehouse for fair comparison. Notably, ConvNeXt-Tiny employs enhanced training strategies (\eg, advanced data augmentation and learning rate schedules), while other architectures follow conventional training strategies. Detailed configurations include: optimizer is SGD with momentum (0.9) and weight decay (1e-4/4e-5) for ResNet/MobileNetV2, and AdamW with weight decay (5e-2) for ConvNeXt-Tiny. The learning rate is initialized at 0.1 for ResNet, 5e-2 for MobileNetV2, and 4e-3 for ConvNeXt-Tiny, decayed via cosine annealing. Batch size is 256 for ResNet and MobileNetV2, and 1024 for ConvNeXt-Tiny. Training epochs are set to 100/150 for ResNet/MobileNetV2, and 300 for ConvNeXt-Tiny. This setup ensures reproducibility and fairness in comparing parameter efficiency and computational overhead across methods.

\noindent {\bf Experiment Result:}
As can be seen from Table~\ref{table:classif_imagenet}, KDNA achieves SOTA accuracy-efficiency trade-offs across diverse architectures (ResNet18/50, MobileNetV2, ConvNeXt-Tiny) while maintaining hardware-friendly inference. Compared to dynamic convolution variants like DY-Conv and ODConv, KDNA reduces parameters by 1.2--5$\times$ (\eg, 9.24M vs. 44.90M on ResNet18) with comparable or lower FLOPs, yet outperforms them in accuracy (\eg, 74.23\% vs. 73.97\% Top-1 on ResNet18). Unlike parameter-heavy methods such as CondConv (81M params), KDNA preserves near-original throughput (7,683 fps vs. ResNet18's 8,463 fps) and minimizes CPU latency (28.43ms vs. ODConv's 54.34ms). FDConv achieves comparable parameter efficiency (11.75M vs. 9.24M on ResNet18) but suffers from severe inference overhead (422 fps, 18$\times$ slower than KDNA) with inferior accuracy (71.25\% vs. 74.23\% Top-1), demonstrating that frequency-domain transformations introduce prohibitive computational costs.

\begin{table}[ht]\small
    \centering
    \caption{Results comparison on ImageNet-1K dataset. The ``MbV2" denotes the ``MobileNetV2", ``Conv-T" denotes the ConvNeXt-Tiny. Pm: Parameters, FP: FLOPs, TP: Throughput, Lt: Latency, Acc: Accuracy. ~\textit{Best results are bolded.}}
    \label{table:classif_imagenet}
    \resizebox{1.0\linewidth}{!}{
    \begin{tabular}{l|r|c|c|c|c|c}
    \toprule
    Model            & \makecell{Pm\\(M)}    & \makecell{FP\\(G)}   &\makecell{TP GPU\\(fps)} &\makecell{Lt CPU\\(ms)}    & \makecell{Top-1\\Acc(\%)}  & \makecell{Top-5\\Acc(\%)}   \\
    \hline
    ResNet18         & 11.69    & 1.82  &    8463    &   22.16    & 70.25          & 89.38 \\
    + SE             & 11.78    & 1.82  &    8319    &   23.68    & 70.98          & 90.03 \\
    + CBAM           & 11.78    & 1.82  &    6023    &   25.25    & 71.01          & 89.85 \\
    + ECA            & 11.69    & 1.82  &    8372    &   23.54    & 70.60          & 89.68 \\
    + CGC            & 11.69    & 2.19  &    1537    &   78.92    & 71.60          & 90.35 \\
    + DCD            & 14.70    & 1.90  &    5305    &   41.43    & 72.33          & 90.65 \\
    + CondConv       & 81.35    & 1.82  &    280     &   89.90    & 71.99          & 90.27 \\
    + DY-Conv        & 45.47    & 1.87  &    3308    &   46.15    & 72.76          & 90.79 \\
    + ODConv         & 44.90    & 1.82  &    1267    &   54.34    & 73.97          & 91.35 \\
    + KW             & 11.93    & 2.23  &    2272    &   69.62    & 73.67          & 91.17 \\
    + FDConv         & 11.75    & 1.92  &    422     &   1013     & 71.25          & 90.04 \\
    + KDNA           & 9.24     & 2.28  &    7683    &   28.43    & {\bf 74.23}    & {\bf 92.38}   \\
    \hline
    ResNet50         & 25.56    & 4.11  &    2432    &   90.77    & 78.44          & 94.24 \\
    + DY-Conv        & 100.88   & 4.22  &    1161    &   155.63   & 79.00          & 94.27  \\
    + ODConv         & 90.67    & 4.12  &    560     &   132.39   & 80.62          & 95.16 \\
    + KW             & 28.05    & 6.15  &    922     &   138.78   & 80.38          & 95.19  \\
    + FDConv         & 27.16    & 4.29  &    230     &   1380.65  & 78.76          & 94.35 \\
    + KDNA           & 22.78    & 4.33  &    2346    &   93.12    & {\bf 80.95}    & {\bf 95.54}   \\
    \hline
    MbV2($0.5\times$) & 1.97    & 0.10  &   12787    &   10.04    & 64.30          & 85.21 \\
    + DY-Conv        & 4.57    & 0.11  &   5818     &   45.59    & 69.05          & 88.37  \\
    + ODConv         & 4.44    & 0.11  &   3761     &   62.07    & {\bf 70.01}    & 89.01   \\
    + KW             & 2.85    & 0.16  &   3284     &   26.43    & 68.29          & 87.93   \\
    + KDNA           & 1.88    & 0.11  &   12461    &   10.27    & 69.32          & {\bf 89.04}   \\
    \hline
    MbV2($1.0\times$) & 3.50    & 0.31  &   6323     &   30.21    & 72.02          & 90.43  \\
    + DY-Conv        & 12.40    & 0.33  &   3667     &   46.93    & 74.94          & 91.83   \\
    + ODConv         & 11.52    & 0.32  &   2226     &   59.51    & 75.42          & 92.18    \\
    + KW             & 5.17     & 0.47  &   2288     &   44.58    & 74.68          & 91.90   \\
    + KDNA           & 3.14     & 0.31  &   6260     &   33.46    & {\bf 75.51}    & {\bf 92.27}   \\
    \hline
    Conv-T           & 28.59    & 4.47  &   1847     &   56.20    & 82.07          & 95.86 \\
    + KW             & 32.99    & 5.61  &   752      &   224.98   & 82.55          & 96.08 \\
    + KDNA           & 24.32    & 4.51  &   1739     &   60.12    & {\bf 82.76}    & {\bf 96.34}   \\
    \bottomrule
    \end{tabular}
    }
\end{table}

KDNA also surpasses semi-filters sharing approaches like KernelWarehouse (KW) in both efficiency and performance. With 61\% parameter budgets, KDNA achieves higher accuracy (75.51\% vs. KW's 74.68\% on MobileNetV2($1.0\times$)) while retaining 99\% of the base model's throughput (6,260 fps vs. 6,323 fps). Its static-dynamic decoupling ensures minimal computational overhead, outperforming KW's kernel assembly (4.33G FLOPs vs. KW's 6.15G FLOPs on ResNet50). On ResNet50, FDConv exhibits comparable parameter efficiency (27.16M vs. 22.78M) but suffers from severe inference slowdown (230 fps, 10.2$\times$ slower than KDNA) with inferior accuracy (78.76\% vs. 80.95\% Top-1), highlighting the importance of maintaining standard convolution's efficient structure. These results validate KDNA's ability to unify parameter efficiency, hardware compatibility, and adaptive feature learning, making it a practical solution for real-world deployment.

\subsection{Detection and Segmentation on MS-COCO}
\noindent {\bf Experiment Setup:}
To further validate the generalization capability of our method, we conduct comparative experiments on downstream tasks, including object detection and instance segmentation, against state-of-the-art dynamic convolution approaches. Specifically, we evaluate on the MS-COCO dataset (118K training and 5K validation images), using Mask R-CNN~\cite{he2017mask} with ImageNet-1K pre-trained models as the backbone. For fairness, all models adhere to the same training strategies as KernelWarehouse~\cite{Li2024}, including identical data preprocessing (\eg, random resizing, flipping) and hyperparameter configurations. The evaluation metrics include Average Precision (AP) and its variants (AP$_{50}$, AP$_{75}$, AP$_S$, AP$_M$, AP$_L$) for object detection and instance segmentation. All experiments are conducted using the MMDetection~\cite{mmdetection} framework with the same evaluation settings.

\noindent {\bf Experiment Result:} Table~\ref{table:coco} compares the performance of our proposed KernelDNA (KDNA) with state-of-the-art dynamic convolution methods on object detection and instance segmentation tasks. Evaluated across diverse backbones (ResNet50, MobileNetV2, ConvNeXt-Tiny), KDNA consistently outperforms existing approaches in multiple metrics, demonstrating superior parameter efficiency. 
\begin{table*}[ht]
    \centering
    \caption{Results comparison on MS-COCO dataset using the pre-trained backbone models.~\textit{Best results are bolded.}}
    \label{table:coco}
    \resizebox{0.9\linewidth}{!}{
    \begin{tabular}{c|l|c|c|c|c|c|c|c|c|c|c|c|c}
    \toprule
    \multirow{2}*{\makecell[c]{Detector}}  & \multirow{2}*{\makecell[c]{Backbone Model}} & \multicolumn{6}{c|}{Object Detection} &  \multicolumn{6}{c}{Instance Segmentation} \\
    \cline{3-14}
    & & $AP $ & $AP_{50} $ & $AP_{75} $ & $AP_{S} $ & $AP_{M} $ & $AP_{L} $ & $AP $ & $AP_{50} $ & $AP_{75} $ & $AP_{S} $ & $AP_{M} $ & $AP_{L} $ \\
    \hline
    \multirow{14}*{\makecell[c]{Mask R-CNN}} 
    & ResNet50       & 39.6       & 61.6       & 43.3       & 24.4       & 43.7       & 50.0       & 36.4       & 58.7       & 38.6       & 20.4       & 40.4       & 48.4 \\
    & + DY-Conv      & 39.6       & 62.1       & 43.1       & 24.7       & 43.3       & 50.5       & 36.6       & 59.1       & 38.6       & 20.9       & 40.2       & 49.1 \\
    & + ODConv       & 42.1       & 65.1       & 46.1       & 27.2       & 46.1       & 53.9       & 38.6       & 61.6       & 41.4       & 23.1       & 42.3       & 52.0 \\
    & + KW           & 41.8       & 64.5       & 45.9       & 26.6       & 45.5       & 53.0       & 38.4       & 61.4       & 41.2       & 22.2       & 42.0       & 51.6 \\
    & + KDNA         & {\bf 42.5} & {\bf 66.2} & {\bf 46.5} & {\bf 28.0} & {\bf 46.5} & {\bf 54.5} & {\bf 39.1} & {\bf 61.9} & {\bf 41.5} & {\bf 23.5} & {\bf 42.8} & {\bf 52.6} \\
    \cline{2-14}
    & MobileNetV2 ($1.0\times$) & 33.8       & 55.2       & 35.8       & 19.7       & 36.5       & 44.4       & 31.7       & 52.4       & 33.3       & 16.4       & 34.4       & 43.7 \\
    & + DY-Conv      & 37.0       & 58.6       & 40.3       & 21.9       & 40.1       & 47.9       & 34.1       & 55.7       & 36.1       & 18.6       & 37.1       & 46.3 \\
    & + ODConv       & 37.2       & 59.4       & 39.9       & 22.6       & 40.0       & 48.0       & 34.5       & 56.4       & 36.3       & 19.3       & 37.3       & 46.8 \\
    & + KW           & 36.4       & 58.3       & 39.2       & 22.0       & 39.6       & 47.0       & 33.7       & 55.1       & 35.7       & 18.9       & 36.7       & 45.6 \\
    & + KDNA         & {\bf 37.3} & {\bf 59.8} & {\bf 40.5} & {\bf 23.0} & {\bf 40.7} & {\bf 48.5} & {\bf 34.8} & {\bf 56.5} & {\bf 36.8} & {\bf 19.5} & {\bf 37.8} & {\bf 47.2} \\
    \cline{2-14}
    & ConvNeXt-Tiny  & 43.4       & 65.8       & 47.7       & 27.6       & 46.8       & 55.9       & 39.7       & 62.6       & 42.4       & 23.1       & 43.1       & 53.7 \\
    & + KW           & 44.8       & 67.7       & 48.9       & 29.8       & 48.3       & 57.3       & 40.6       & 64.4       & 43.4       & 24.7       & 44.1       & 54.8 \\
    & + FDConv       & 45.2       & 67.9       & 48.8       & 29.9       & 48.7       & 57.6       & 40.8       & 64.8       & 43.6       & 24.8       & 44.5       & 54.9 \\
    & + KDNA         & {\bf 45.4} & {\bf 68.1} & {\bf 49.4} & {\bf 30.4} & {\bf 48.9} & {\bf 58.2} & {\bf 41.2} & {\bf 65.3} & {\bf 43.8} & {\bf 24.9} & {\bf 44.7} & {\bf 55.4} \\
    \bottomrule
    \end{tabular}
    }
    \vspace{-0.5cm}
\end{table*}

Notably, KDNA surpasses 4$\times$-parameter methods like ODConv (detection $AP$: 42.5 vs. 42.1 on ResNet50) and outperforms KernelWarehouse (KW) under equal parameter constraints (detection $AP$: 37.3 vs. 36.4 on MobileNetV2). On ConvNeXt-Tiny, FDConv achieves competitive results (detection $AP$: 45.2, segmentation $AP$: 40.8), but KDNA consistently outperforms FDConv across all metrics (detection $AP$: 45.4 vs. 45.2, segmentation $AP$: 41.2 vs. 40.8), particularly excelling in small-object detection (${AP}_S$: 30.4 vs. 29.9) and large-object segmentation (${AP}_L$: 58.2 vs. 57.6). Crucially, KDNA maintains hardware efficiency via static attention pre-fusion, while FDConv's frequency-domain operations incur significant computational overhead. The consistent superiority of KDNA across all backbones demonstrates its superior balance of accuracy, parameter efficiency, and practicality for real-world vision tasks.

\subsection{Ablation Study}
To comprehensively validate the advantages of our proposed dynamic kernel-sharing mechanism over standard convolution and the effectiveness of the three visual attention modules, we conduct a series of ablation experiments on the ImageNet-1K dataset. 

\noindent {\bf Settings of ``parent" and ``child" convolutional layers.}
Although previous experiments have demonstrated the effectiveness of our proposed dynamic kernel-sharing mechanism, the optimal number and placement of shared kernels (``child") and base kernels (``parent") require further exploration. Let the shared kernels be denoted as ``S" and the base kernels (full kernels) as ``F". Subscripts of ``S" and ``F" indicate parent-child relationships. Considering that convolutional layers within the same stage (typically with consistent kernel sizes) are more amenable to parameter sharing, our experiments focus on sharing and extending kernels across all blocks within the same stage, as is shown in Table~\ref{table:ablation_SF_relation}.

\begin{table}[ht]
    \centering
    \caption{Permutation of ``parent" and ``child" Conv layers. }
    \label{table:ablation_SF_relation}
    \resizebox{1.0\linewidth}{!}{
    \begin{tabular}{l|c|c|c|c|c}
    \toprule
    Model       & Parent-Child  & Params  &\makecell{TP GPU\\(fps)}  & \makecell{Top-1\\Acc(\%)} & \makecell{Top-5\\Acc(\%)} \\
    \hline
    \multirow{6}*{\makecell[c]{ResNet18}}   
                & FF-FF(orig)  & 11.69M  &   8463    &  70.25    & 89.38 \\
                & FF-FFF           & 14.64M  &   7944    &  73.41    & 91.37  \\
                & FF-FF-FF         & 17.59M  &   7496    &  74.07    & 92.23 \\
                & FS-FSF           & 9.07M   &   7814    &  73.95    & 92.61 \\
                & FS-SSF           & 6.29M   &   7756    &  73.42    & 91.64  \\
                & FS-SF-SF         & 9.24M   &   7683    &  {\bf 74.23}    & {\bf 92.38}  \\
    \bottomrule
    \end{tabular}
    }
    \vspace{-0.5cm}
\end{table}

From the results, it is evident that the placement of parent and child kernels is crucial for model performance. Using preceding layers as child kernels for a parent layer yields better results than using subsequent layers. Additionally, as the number of shared child kernels increases, the model's performance gains diminish and may even degrade. Thus, simply scaling up the number of child kernels has an upper limit. However, scaling up the number of parent kernels gradually improves performance, albeit at the cost of increased parameters. Based on these experiments, we find that adopting the ``FS-SF-SF" sequence and quantity within the same block achieves the best trade-off between inference efficiency and performance. Note, the ``-" indicates the position for the residual summation, ``orig" indicates the original sequence. 

\noindent {\bf Comparison of methods to increase convolutional layers.}
To validate the advantages of shared child kernels generated from parent kernels via lightweight adapters over simply adding more convolutional layers or directly copying parent kernels, we conduct experiments with three approaches:
\begin{itemize}
    \item {\bf Expanding Parent Kernels (Expand):} Adding more static convolutional layers.
    \item {\bf Direct Copying (Copy):} Creating shared child kernels by directly duplicating parent kernels.
    \item {\bf Adapter-Based Generation (Adapter):} Producing shared child kernels through lightweight adapters (Ours).
\end{itemize}

\begin{table}[ht]
    \centering
    \caption{Comparison of methods for increasing Conv layers.}
    \label{table:ablation_Increase_layers}
    \resizebox{0.95\linewidth}{!}{
    \begin{tabular}{l|c|c|c|c|c}
    \toprule
    Model       & \makecell{Increase\\layers}  & Params  &\makecell{TP GPU\\(fps)}  & \makecell{Top-1\\Acc(\%)} & \makecell{Top-5\\Acc(\%)} \\
    \hline
    \multirow{4}*{\makecell[c]{ResNet18}}   
                & Orig       & 11.69M  &   8463    &  70.25    & 89.38   \\
                & Expand     & 14.64M  &   7744    &  73.41    & 91.37   \\
                & Copy       & 8.74M   &   7735    &  69.85    & 89.41   \\
                & Adapter    & 9.24M   &   7683    &  {\bf 74.23}    & {\bf 92.38}   \\
    \hline
    \multirow{4}*{\makecell[c]{ResNet50}}   
                & Orig       & 25.56M  &   2432    &  78.44    & 94.24   \\
                & Expand     & 26.67M  &   2358    &  80.23    & 94.83    \\
                & Copy       & 22.55M  &   2355    &  75.21    & 92.87    \\
                & Adapter    & 22.78M  &   2346    &  {\bf 80.95}    & {\bf 95.54}   \\
    \bottomrule
    \end{tabular}
    }
    \vspace{-0.5cm}
\end{table}

The results are presented in Table~\ref{table:ablation_Increase_layers}. From the experimental outcomes, it is evident that our adapter-based approach achieves a superior trade-off between performance and efficiency compared to the other two methods.

\noindent {\bf Analysis of the three attention components.}
Our comparative study on different methods of expanding convolutional layers highlights the importance of parameter sharing through a ``bridge" (i.e., the adapter), and the design of this bridge is equally crucial. In this work, we decouple visual attention mechanisms across dimensions into dynamic and static components, addressing channel-wise, spatial-wise, and filter-wise kernel modulation. Compared to previous works like ODConv, which employs additional dynamic and sequential kernel attention, our implementation is simpler and more efficient.

\begin{table}[ht] \small
    \centering
    \caption{Ablation of the three attention modules (ResNet18).}
    \label{tab:abla_attens}
    \resizebox{1.0\linewidth}{!}{
    \begin{tabular}{c|c|c|c|c|c|c}
    \toprule
    \makecell{Channel} & \makecell{Spatial} & \makecell{Filter} & Params  &\makecell{TP GPU\\(fps)}  & \makecell{Top-1\\Acc(\%)} & \makecell{Top-5\\Acc(\%)} \\
    \hline   
     \ding{55}  & \ding{55}  & \ding{55}   & 8.74M   &  7863  &  69.85    & 89.41  \\
     \ding{51}  & \ding{55}  & \ding{55}   & 9.24M   &  7726  &  71.82    & 90.72  \\
     \ding{55}  & \ding{51}  & \ding{55}   & 8.74M   &  7815  &  70.90    & 90.06  \\
     \ding{55}  & \ding{55}  & \ding{51}   & 8.75M   &  7839  &  70.29    & 89.29  \\
     \ding{51}  & \ding{51}  & \ding{55}   & 9.24M   &  7698  &  72.68    & 91.54  \\
     \ding{51}  & \ding{55}  & \ding{51}   & 9.24M   &  7699  &  71.98    & 91.05  \\
     \ding{55}  & \ding{51}  & \ding{51}   & 8.75M   &  7804  &  71.01    & 90.23  \\
     \ding{51}  & \ding{51}  & \ding{51}   & 9.24M   &  7683  &  {\bf 74.23}    & {\bf 92.38}  \\
    \bottomrule
    \end{tabular}
    }
    \vspace{-0.3cm}
\end{table}

To further validate the effectiveness of the three attention components, we conduct ablation studies (see Table~\ref{tab:abla_attens}). The results show that all three attention mechanisms contribute positively to performance. Notably, Channel Attention delivers the most significant accuracy gains, albeit with a slight impact on inference efficiency. In contrast, Spatial Attention and Filter Attention achieve moderate performance improvements without introducing additional computational overhead. These findings collectively confirm the efficacy of our lightweight adapter's tripartite attention design.

\subsection{Parameter Scaling}
KDNA also supports flexible parameter scaling by adjusting the number of parent and child kernels. Table~\ref{table:scaling} presents the results of scaling KDNA on both ResNet18 and MobileNetV2($0.5\times$). As the scale increases, KDNA consistently improves accuracy with a corresponding increase in parameter count, demonstrating its scalability across different model sizes. 
\begin{table}[ht]
    \centering
    \caption{Results of parameter scaling on ImageNet-1K.}
    \label{table:scaling}
    \resizebox{0.95\linewidth}{!}{
    \begin{tabular}{llccc}
    \toprule
    Model  & Scale               & Params    & \makecell{Top-1\\Acc(\%)}  & \makecell{Top-5\\Acc(\%)}   \\
    \midrule
    \multirow{4}{*}{ResNet18}
          &baseline              & 11.69M    & 70.25      & 89.38 \\
          &KDNA (1$\times$)      & 9.24M     & 74.23      & 92.38 \\
          &KDNA (2$\times$)      & 18.48M    & 75.41      & 92.75 \\
          &KDNA (4$\times$)      & 45.91M    & 76.58      & 93.26 \\
    \midrule
    \multirow{4}{*}{MobileNetV2($0.5\times$)}
          &baseline              &  1.97M    & 64.30      & 85.21  \\
          &KDNA (1$\times$)      &  1.88M    & 69.32      & 89.04  \\
          &KDNA (2$\times$)      &  3.24M    & 70.82      & 89.73  \\
          &KDNA (4$\times$)      &  4.46M    & 71.36      & 90.22  \\
    \bottomrule
    \end{tabular}
    }
    \vspace{-0.5cm}
\end{table}

\section{Discussion}
\subsection{Why Kernel Sharing is Effective?}
While experiments validate the effectiveness of KernelDNA across multiple vision tasks, further exploration into its underlying mechanisms is critical. We conduct Linear CKA analysis on ResNet18 trained with KernelDNA under optimal configurations on ImageNet-1K (Figure~\ref{fig:compare_CKA}). The left subplot shows the similarity grid of the original ResNet18 layers, while the middle subplot represents KernelDNA's results, where parameter-sharing replaces or extends original convolutional layers, resulting in slightly more layers. The right subplot compares layer-wise similarity between KernelDNA and the original model. We observe that:
\begin{figure}[ht]
    \centering
    \includegraphics[width=0.47\textwidth]{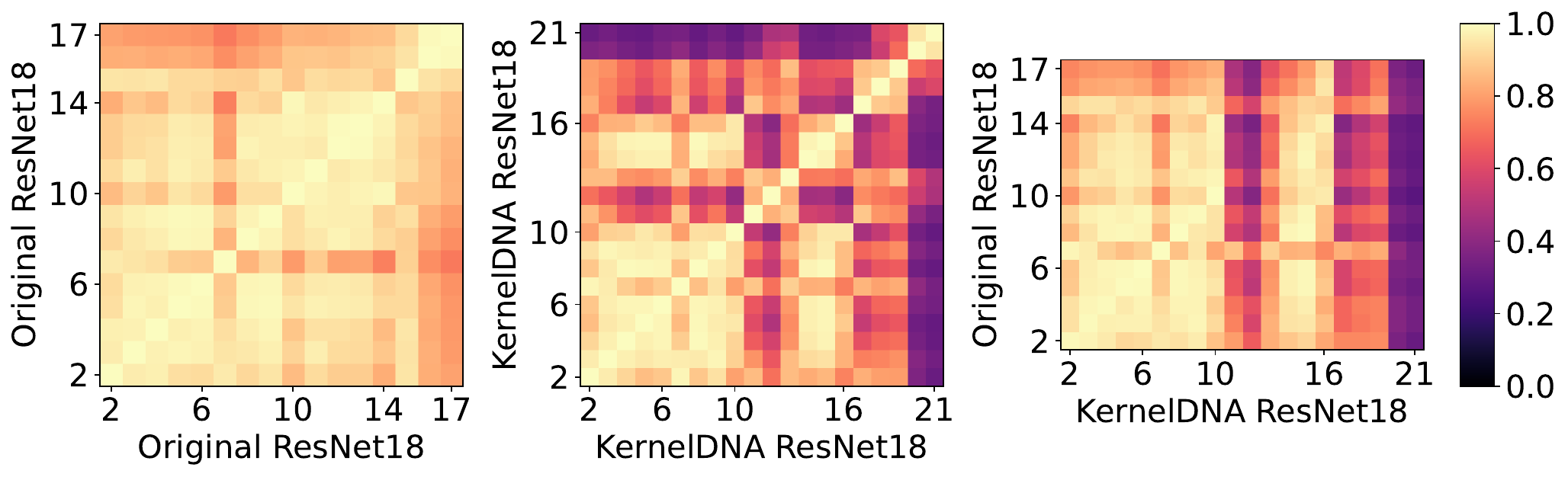}
    \caption{Linear CKA between the original model and the model with KernelDNA, only the results for Conv3$\times$3 are presented.}
    \label{fig:compare_CKA}
\end{figure}

\noindent {\bf Preservation of Structural Patterns:} KernelDNA maintains the original model's global similarity distribution—the last two layers and a few intermediate layers exhibit lower similarity (indicating task-specific representation learning), while other layers remain highly similar.

\noindent {\bf Block-Wise Specificity:} KernelDNA retains intra-block commonalities (same parent) between parent and child layers while introducing inter-block heterogeneity (different parents), aligning with principles of epigenetic regulation in feature learning.

\noindent {\bf Parameter Sharing Validity:} High similarity between specific layers of KernelDNA and the original model confirms that shared parameters preserve critical learned features, demonstrating the efficacy of our mechanism.

This analysis bridges biological inspiration with neural architecture design, showing how KernelDNA balances redundancy reduction with task-specific adaptability, offering a principled framework for efficient dynamic networks.

\subsection{Do Adapters Really Learn the Differences?}
\begin{figure}[ht]
    \centering
    \includegraphics[width=0.47\textwidth]{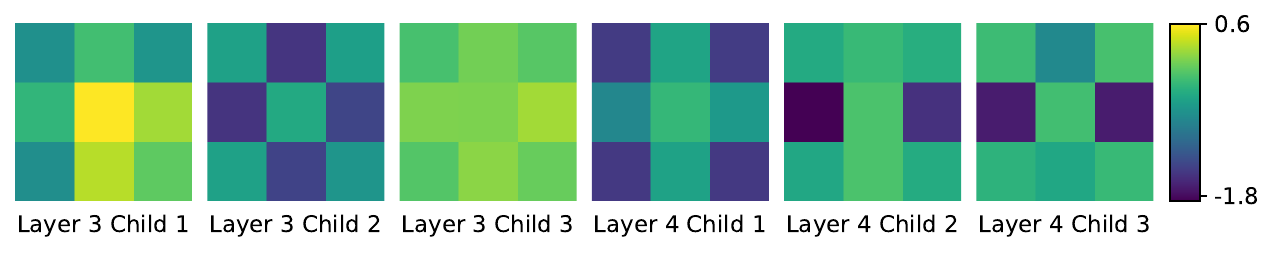}
    \caption{Visualization of adapter learning differences between parent and child kernels in ResNet18.}
    \label{fig:adapter_learn_diff}
\end{figure}

To further validate whether our proposed ``DNA" adapter—equipped with attention mechanisms—genuinely learns diversity across shared child layers, we visualize the spatial attention features, as shown in Figure~\ref{fig:adapter_learn_diff}. The figure compares three child layers derived from the third parent layer and three from the fourth parent layer. The results demonstrate that child layers from the same parent exhibit distinct variations, confirming that our adapter preserves shared commonality features from parent kernels while introducing differentiated diversity among child layers. This aligns with our hypothesis that the adapter balances inheritance and specialization, ensuring both parameter efficiency and expressive power.

\section{Conclusion}
KernelDNA introduces a lightweight convolution plugin that addresses the parameter and efficiency limitations of dynamic convolutions. By leveraging cross-layer weight sharing and adapter-based modulation, it combines dynamic kernel adaptation with standard convolution's efficiency, reducing parameters while enhancing performance. Experiments show state-of-the-art (SOTA) accuracy-efficiency trade-offs on image classification and dense prediction tasks, outperforming existing methods. KernelDNA maintains hardware-friendly inference speeds with 1.2--5$\times$ fewer parameters than previous SOTA variants, offering a practical solution for adaptive CNNs with accuracy profitability.

{
    \small
    \bibliographystyle{ieeenat_fullname}
    \bibliography{main}
}

\clearpage
\setcounter{page}{1}
\maketitlesupplementary

\setcounter{section}{0}
\renewcommand{\thesection}{\Alph{section}}
\renewcommand{\thesubsection}{\Alph{section}.\arabic{subsection}}
\renewcommand{\thesubsubsection}{\Alph{section}.\arabic{subsection}.\arabic{subsubsection}}

\section{Limitations}
Despite the substantial parameter reduction achieved by our kernel-sharing strategy, there are some limitations to consider. KernelDNA may introduce a slight increase in computational overhead (e.g., marginally higher FLOPs) due to the adapter modules, although our experiments demonstrate that the impact on inference speed is minimal—maintaining 90--99\% throughput of the base models. This trade-off may be less suitable for extremely compute-constrained environments. However, given that most modern hardware is memory-bound rather than compute-bound, the parameter efficiency and hardware-friendly design of KernelDNA makes it well-suited for a wide range of real-world applications.

\section{Additional Experiments} 
\subsection{Channel Reduction and Expansion}
To further investigate the flexibility of KDNA, we conduct experiments with various channel reduction and expansion ratios, as summarized in Table~\ref{table:channel_reduction}. The results demonstrate that the default configuration (highlighted in bold) achieves the optimal balance between parameter efficiency and inference speed, with only marginal accuracy changes observed as the ratio varies.

\begin{table}[ht]
  \centering
  \caption{Results of different channel reduction ratios on ResNet18.}
  \label{table:channel_reduction}
  \resizebox{0.95\linewidth}{!}{
  \begin{tabular}{lccccc}
  \toprule
  Type  & Ratio              & Params     &\makecell{TP GPU\\(fps)}  & \makecell{Top-1 \\Acc(\%)}  & \makecell{Top-5 \\Acc(\%)}   \\
  \midrule
  \multirow{5}{*}{Channel}
        &1                   &  10.719M   & 7670  & 74.28      & 92.42 \\
        &1/2                 &  9.733M    & 7678  & 74.25      & 92.41 \\
        &{\bf 1/4}           &  9.241M    & 7683  & {\bf 74.23}      & {\bf 92.38} \\
        &1/8                 &  8.994M    & 7688  & 74.16      & 92.26 \\
        &1/16                &  8.871M    & 7691  & 73.76      & 91.91 \\
  \midrule
  \multirow{5}{*}{Spatial}
        &{\bf 1}             &  9.241M    & 7683  & {\bf 74.23}      & {\bf 92.38} \\
        &2                   &  9.241M    & 7682  & 74.22      & 92.36 \\
        &4                   &  9.241M    & 7678  & 74.19      & 92.29 \\
        &8                   &  9.241M    & 7670  & 74.21      & 92.30 \\
        &16                  &  9.242M    & 7665  & 74.11      & 92.23 \\
  \midrule
  \multirow{5}{*}{Filter}
        &{\bf 1}             &  9.241M    & 7683  & {\bf 74.23}      & {\bf 92.38}  \\
        &2                   &  9.243M    & 7679  & 74.21      & 92.35  \\
        &4                   &  9.248M    & 7675  & 74.18      & 92.29  \\
        &8                   &  9.257M    & 7671  & 74.19      & 92.31  \\
        &16                  &  9.275M    & 7662  & 74.12      & 92.26  \\
  \bottomrule
  \end{tabular}
  }
\end{table}


\end{document}